\documentclass{article} 
\usepackage{colm2024_conference}

\usepackage{microtype}
\usepackage{hyperref}
\usepackage{url}
\usepackage{booktabs}
\usepackage{graphicx}
\usepackage{amsmath}
\usepackage{colortbl}
\usepackage{multirow}

\title{GPTA: Generative Prompt Tuning Assistant for Synergistic Downstream Neural Network Enhancement with LLMs}

\author{Xiao Liu, Jiawei Zhang \\
IFM Lab, Department of Computer Science\\
University of California, Davis\\
\texttt{\{xioliu, jiwzhang\}@ucdavis.edu} \\
}

\colmfinalcopy 
\begin{document}

\maketitle

\begin{abstract}

This study introduces GPTA, a Large Language Model assistance training framework, that enhances the training of downstream task models via prefix prompt. By minimizing data exposure to LLM, the framework addresses the security and legal challenges of applying LLM in downstream task model training. GPTA utilizes a new synergistic training approach, optimizing the downstream models with parameter gradients and LLMs with the novel ``dialogue gradient''. The framework not only demonstrates significant improvements in model performance across six NLP benchmark datasets, but also reduces overfitting in low-resource scenarios effectively. The detailed analyses further validate that our pioneer framework provides a cost-efficient and adaptive method for downstream task model training with LLM support.
\end{abstract}

\section{Introduction}

Large Language Models (LLMs) have achieved remarkable success in a variety of open-domain tasks, including sentiment analysis, machine reading comprehension, and text summarization \citep{li2023reinforcement, openai2023gpt4, bang-etal-2023-multitask}. 
Presently, LLMs are broadly classified into two categories: open-sourced models, exemplified by Llama2 \citep{touvron2023llama}and Gemma \citep{googleGemma2023}, and API-based models, such as ChatGPT \citep{openai2023gpt4} and Claude 3 \citep{claude2023}. The advancements in LLM research and development have seamlessly integrated these models into numerous aspects of daily life as a powerful tool. 

Nevertheless, recent studies have highlighted that while LLMs perform well in general tasks, they face notable limitations in specialized fields such as medicine, law, and science \citep{lu2023human, luo2023chatgpt}. Moreover, the use and optimization of open-sourced LLMs in those specialized fields presents significant challenges for many enterprises and research institutions. These challenges primarily stem from the high costs related to computational resources and manpower needed for training or deploying these models. Besides, the risks of producing unstable and biased outcomes during LLM training further complicate the challenge \citep{hoffmann2022training}. 


\begin{figure}[ht]
\begin{center}
\includegraphics[width=0.83\textwidth]{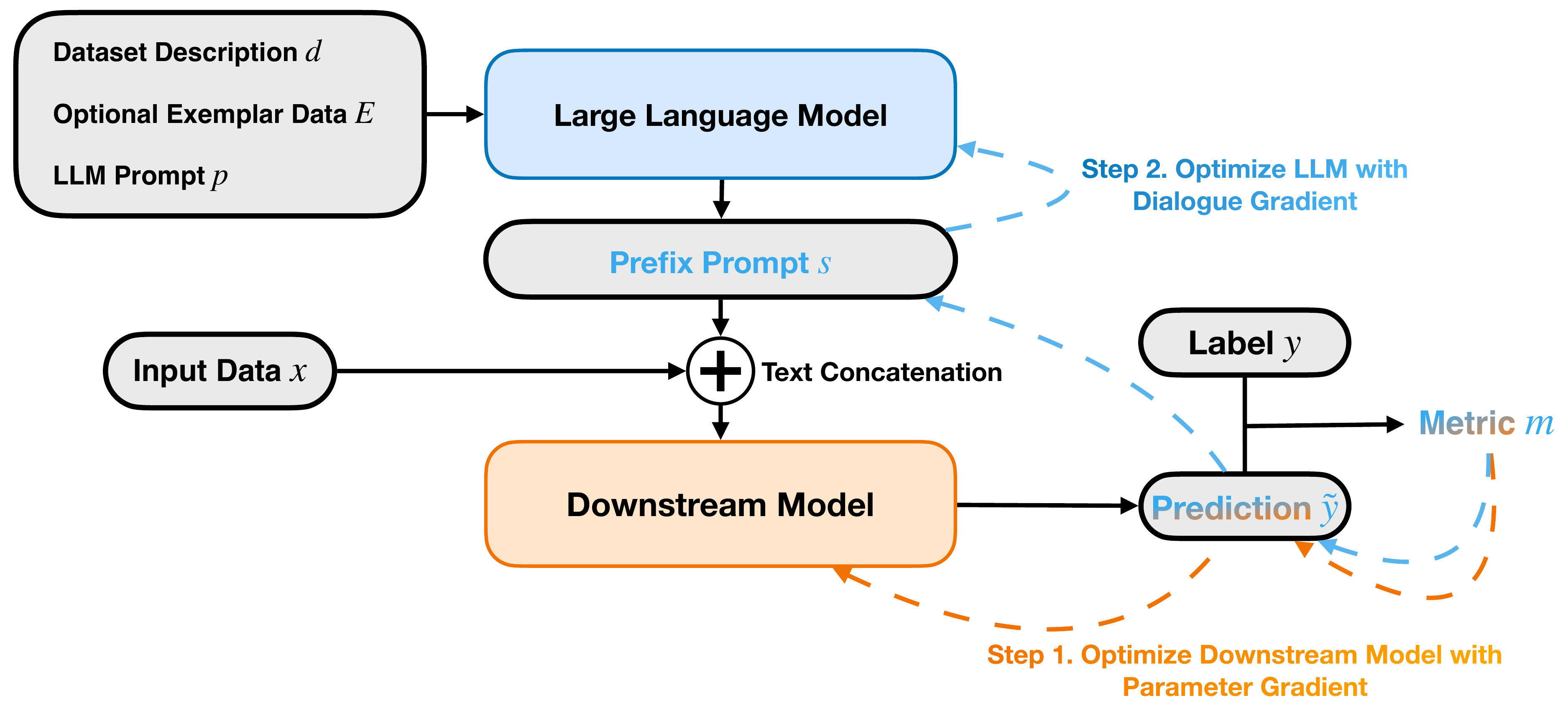}
\end{center}
\caption{Forward and backward process of GPTA. Forward process (solid arrows) involves the LLM generating prefix prompts for downstream model input enhancement. The backward process (dotted arrows) uses gradient descent for downstream optimization, then applies ``dialogue gradient'' for the LLM. Colored texts indicate variables have gradient.}
\label{fig:model_archi}
\end{figure}

In the current transitional period, as we await further advancements in computational capabilities for broader LLM deployment, training models for downstream tasks in those specialized domains remain a viable strategy. Recent research has explored the use of API-based LLMs to facilitate training of downstream models through knowledge distillation \citep{peris2022knowledge, gu2023knowledge, udagawa-etal-2023-comparative, zhu2023survey} and data augmentation \citep{kaddour2023text, edwards-etal-2022-guiding, yang2023neural, sahu-etal-2022-data}. Although these methods can yield high performance with downstream models that are significantly more parameter-efficient, they also introduce concerns regarding data security and legal challenges. Internally, sharing private data with third-party API providers may risk data breaches. Externally, legal constraints imposed by API providers on the use of synthesized data for training commercial models present challenges. Moreover, LLMs employed in these methods are frozen, potentially misunderstanding domain-specific tasks and leading to hallucination and bias issues, thereby misleading downstream models \citep{wang2023grammar, yao2023knowledge, huang2023survey}. 

To address these challenges, we introduce GPTA, a novel training framework that harnesses the power of API-based LLMs and its fine-tuning functionality to assist in the training of downstream models. Note that our approach fundamentally diverges from distillation techniques. Unlike distillation approaches that treat LLMs as ``teacher", our framework adopts a novel perspective by considering the LLMs as ``teaching assistant (TA)". This shift underscores our strategy of utilizing the LLMs to assist, rather than direct, the training of the ``student" downstream models.

In our framework, we utilize API-based LLMs as prefix prompt generators that search for the prefix prompt solely based on the dataset description and optional exemplar data. These prefix prompts are then prepended to the training data serving as auxiliary context. This addition aims to enhance the learning process throughout the training phase of the student model. Additionally, to make LLMs better understand the assistance task and the specific domain, we introduce a more dynamic approach than the conventional frozen model setting. 

Specifically, GPTA incorporates a joint optimization of the TA LLM and the student model through the synergistic training steps, facilitating continuous improvements and adaptations of LLM's knowledge for the task domain and the student model. When training, both models will be tuned with a unified objective to improve the performance of the student model. For student models, we still fine-tune the prediction results with conventional gradient descent optimizers to update the model parameters. Since API-based LLMs could only be optimized with the API provider-required dialogue histories, we improve their ability to find high-quality prefix prompts with a novel ``dialogue gradient" optimization proposed in this paper, which relies on the history of prefix prompts and their performance scores. The forward training and backward optimization processes are both illustrated in Figure \ref{fig:model_archi}.

By treating LLMs solely as prefix prompt generators, very little or almost no training data will be required to prompt LLMs during the entire process, protecting data privacy. Once the training is finished, GPTA will enable downstream models to infer independently, relying solely on original, rather than synthesized data to sidestep legal concerns. Our comprehensive experimental evaluations across six benchmark datasets in three key NLP domains illustrate GPTA's ability to improve model performance and reduce overfitting in resource-scarce scenarios significantly. Our detailed analyses also demonstrate that the LLM's accuracy in locating the next prefix prompts strongly relates to the performance improvement of the downstream model, which further validate the effectiveness of our framework. Additionally, we demonstrate the transferability of optimized prefixes across different datasets within the same task domain.

Our contributions are summarized as follows:
\begin{itemize}
    \item We propose GPTA, a novel training framework that utilize API-based LLM to assist downstream model training. This framework significantly enhances downstream model performance while effectively avoiding current data safety and legal challenges associated with LLM applications.
    \item We bridge the gap of joint training the API-based LLM and the student model toward the same objective with a novel synergistic training paradigm and dialogue gradient. To the best of our knowledge, this is the first attempt aligns the training objectives of downstream neural networks and API-based LLMs.
    \item We conduct comprehensive experiments across six datasets spanning three critical NLP domains, with empirical results validating our framework's effectiveness in diverse resource scenarios. Additionally,  we perform a detailed analysis investigating the effectiveness of the TA LLM utilized during the training process.
\end{itemize}

\section{Related Work}

\subsection{Integrating LLM into Model Training}
Large Language Models have shown exceptional performance on various NLP tasks \citep{li2023reinforcement, openai2023gpt4, bang-etal-2023-multitask, liu2023beyond}. However, their extensive parameter size limits widespread deployment. To address this problem, research has explored integrating LLMs into downstream model training through various approaches. One such method is knowledge distillation \citep{peris2022knowledge, gu2023knowledge, udagawa-etal-2023-comparative, zhu2023survey}, where LLMs first infer from training data, and then downstream models are trained to emulate the LLM's task-specific behavior. Although this significantly reduces parameter size while maintaining performance, it necessitates providing data to the LLM. Given that many high-performing LLMs operate by third-party enterprises through APIs, this method poses a considerable risk of data breaches.

Another line of research aims to enhance existing datasets by prompting LLMs to generate synthetic data, which is then used for model training \citep{kaddour2023text, edwards-etal-2022-guiding, yang2023neural, sahu-etal-2022-data}. This strategy can improve performance but faces legal restrictions from API providers in commercial contexts. Moreover, the quality of synthesized data often lacks stability for neural network training, compared to original datasets \citep{Dewi2021Yolo, Zhdanov2023Automatic}.

Aside from data privacy and legal concerns, both strategies rely on utilizing frozen LLMs. This may limit generalizability to domain-specific areas, potentially misleading downstream models \citep{huang2023survey}.

Our proposed GPTA framework tackles these challenges by leveraging LLMs as adaptive teaching assistants during model training. It employs prefix prompts derived solely from data descriptions and optional exemplars, avoiding direct data exposure. Additionally, the LLM itself is optimized to adapt to the domain and update alongside the student model, ensuring domain relevance and mitigating the risks associated with frozen LLM applications.

\subsection{Prompt Tuning}
Recent studies have demonstrated that LLMs are sensitive to variations in input prompts, even when these prompts convey identical semantic meanings \citep{Loya2023Exploring, chen2023unleashing}. This observation has catalyzed the research process on prompt tuning. Further research reveals that LLM performance can be significantly enhanced by the simple addition of a prefix prompt, such as ``Let's take a deep breath" \citep{raffel2023exploring, cheng2023blackbox}. However, these investigations have been limited to the application of prompt tuning methods on frozen LLMs.

Building on this foundational work, our GPTA framework incorporates the concept of a prefix prompt into the training process of smaller-sized models with the assistance of a prefix prompt generator LLM.

\section{Method}
\label{sec:method}
\subsection{The GPTA Framework}
\label{sec:framework}

Our inspiration derives from \cite{yang2023large}'s pioneering study, which demonstrates the remarkable potential of enhancing LLM's performance through the integration of simple prefix prompts like ``Let's take a deep breath" into the input. Leveraging this discovery, we aim to adapt the concept of prefix prompts for the training of smaller models, rather than limiting their use to frozen LLMs. Our GPTA framework utilizes LLMs as dynamic prefix prompt generators, responsible for finding prompts that substantially enhance learning effectiveness in downstream model training.

As shown in Figure \ref{fig:model_archi}, the GPTA framework incorporates two principal components during the training phase:

\begin{enumerate}
    \item \textbf{Downstream Task Model (Student):} This component is designed to learn and perform the specific downstream task. The model is flexible in its architecture, allowing for various parameter-trainable neural networks that process text input. 

    
    \item \textbf{Large Language Model (Teaching Assistant):} Contrary to traditional roles in knowledge distillation where models may act as a "teacher", in our framework, we regard the LLM as the "teaching assistant (TA)". It is instructed to generate prefix prompts based on data description and optional few-shot examples. These prefix prompts are utilized to guide the downstream model's learning process from the dataset. Notably, the involvement of the LLMs ends with the training phase, eliminating its necessity during inference.
\end{enumerate}

Formally, given a supervised text dataset with $n$ text-label pairs as $D = \{(x_1, y_1), (x_2, y_2), ..., (x_n, y_n)\}$, where $x_i$ is the $i$-th  input text and $y_i$ is the ground-truth label. The LLM is requested to generate a prefix prompt $s$ conditioned on the dataset description $d$, exemplar data samples $E$, and prompt $p$:
\begin{equation} 
    s = \operatorname{argmax}_s P_{LLM}(s|p, d, E),
\end{equation}
where $E \subset D$ and $|E| \ll |D|$. $P_{LLM}$ is the conditional probability distribution of $s$ generated by LLM. Note that $E$ is optional for the prefix prompt generation.

Once the prefix prompt is acquired, we first prepend the prefix prompt at the beginning of each input text $x_i$ and send the prefix prompt enhanced input to the downstream model to get the model prediction $\Tilde{y}_i$:

\begin{equation}
\Tilde{y}_i = f_{\theta}([s, x_i])
\label{equ:d_model}
\end{equation}

Here, $f_{\theta}(\cdot)$ represents the downstream model's prediction under the current model parameters $\theta$. Notation $[\cdot, \cdot]$ denotes represents the text concatenation operation.

\subsection{Synergistic Training Models in GPTA}

One of the primary challenges in optimizing API-based LLMs lies in their inherent design, which typically only allows for optimization based on language modeling objectives, using dialogue history as input \footnote{\url{https://platform.openai.com/docs/guides/fine-tuning}}. Those optimizations are often employed to adjust the tone of text or to structure LLM outputs, rather than to execute specific tasks. To address this obstacle, we develop a strategy that integrates optimization directions into the dialogue history itself. By embedding the target optimization objectives within the dialogue history, it becomes possible to direct API-based LLMs toward the desired behavior through the language modeling objective. We term this technique ``dialogue gradient", where the objective-injected dialogue history facilitates targeted optimization of API-based LLMs. 

To accommodate the GPTA framework for API-based LLMs, which cannot be directly optimized with parametric gradients in conjunction with the student model, we introduce a new synergistic joint training process. More specifically, our methodology seeks to optimize the student model and the TA model alternately, but in a unified direction during the training phase.

The training process consists of four major steps in one training epoch: 1)  Downstream Model Training, 2) Prefix Prompt History Collection, 3)  Dialogue Gradient Computation, and 4) LLM Optimization. We will introduce these four steps in detail as follows.

\subsection{Downstream Model Training}

We initiate the training of our downstream model to facilitate its learning of the basic mapping between the input and output. Initially, the LLM is prompted to generate the first prefix prompt, $s_0$. This prefix prompt is employed in conjunction with Equation \ref{equ:d_model} to derive the prediction, $\Tilde{y}_0 = f_{\theta}([s_0, x])$.



Then the loss $\mathcal{L}(y, \Tilde{y}_0)$ is computed as the difference between the ground-truth label $y$ and the predicted output $\Tilde{y}_0$.

Finally, the parameters $\theta$ of the model are updated through gradient descent, where $\alpha$ denotes the learning rate, and $\nabla_{\theta} \mathcal{L}(y, \Tilde{y}_0)$ signifies the gradient of the loss with respect to the model parameters:
\begin{equation}
\theta_{\text{new}} = \theta - \alpha \nabla_{\theta} \mathcal{L}(y, \Tilde{y}_0).
\end{equation}

Notably, we freeze the downstream model when this training step ends.

\subsection{Prefix Prompt History Collection}
Inspired by \citet{yang2023large}, we leverage the in-context learning capabilities of LLMs \citep{dong2023survey}, to assemble a series of $k$ prefix prompt-metric pairs in ascending order, denoted as $H=\{(s_0, m_0), (s_1, m_1), ..., (s_k, m_k)\}$. This preparatory step is crucial for the subsequent computation of dialogue gradients, with the process illustrated in Figure \ref{fig:dialogue_gradient}(a).

At any given timestep $t$, given the current prefix prompt history, $H_t$, consists of $j$ such pairs, where $0 < j < k$. To augment this collection, we engage the LLM to generate a sequence of $l$ prefix prompts which could potentially enhance the performance of the downstream model: 
\begin{equation}
\{s_{j+1}, s_{j+2}, ..., s_{j+l}\} = \operatorname{argmax}_s P_{LLM}(s|p, d, E, H_t).
\end{equation}

Subsequently, for each new prefix prompt $s_n$, we concatenate it with the input data and obtain predictions using the downstream model, which remains frozen during this process. These predictions are then evaluated by a predefined metric function, $\operatorname{metric}(\cdot)$:
\begin{equation}
\Tilde{y}_n = f_{\theta}([s_n, x]), \ \ m_n = \operatorname{metric}(y, \Tilde{y}_n),
\end{equation}
where $\operatorname{metric}(\cdot)$ may align with the training loss function or other suitable text-based automatic evaluation metrics.

The newly generated $l$ prefix prompt-metric pairs are appended to $H_t$. Following the insights proposed by \citet{yang2023large},  the expanded history is sorted in ascending order by the metric scores to derive $H_{t+1}$:
\begin{equation}
H_{t+1} = \operatorname{sort}(H_t \cup \{(s_{j+1}, m_{j+1}), (s_{j+2}, m_{j+2}), ..., (s_{j+l}, m_{j+l})\}).
\end{equation}
This iterative process continues until the history encompasses the desired amount of prefix prompt-metric pairs. The LLM prompt example is shown in Table \ref{tab:prefix_generation} in Appendix \ref{app: llm_prompts}.

\subsection{Dialogue Gradient Computation and LLM Optimization}
\begin{figure}[!t]
\begin{center}
\includegraphics[width=0.90\textwidth]{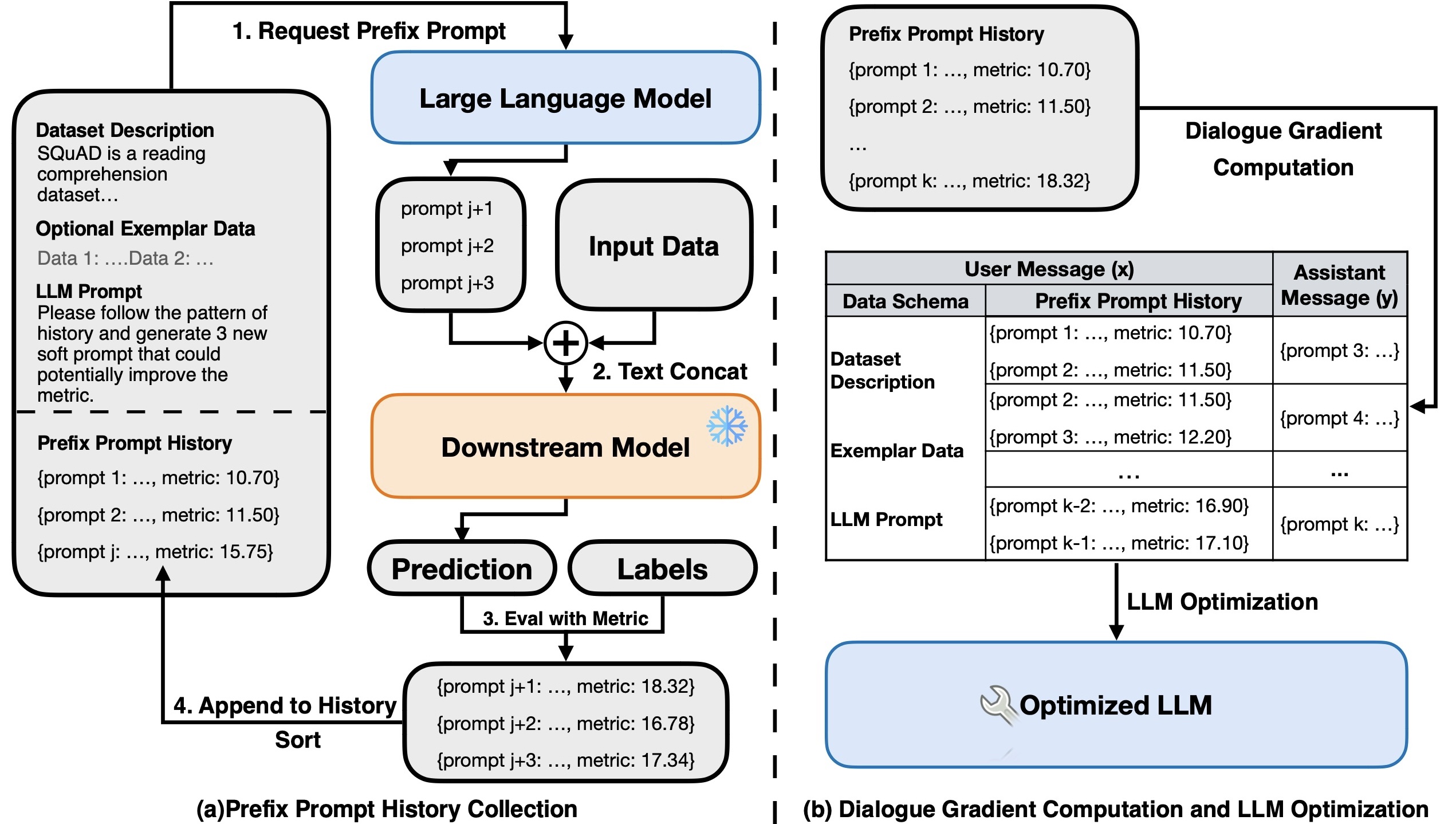}
\end{center}
\caption{Prefix Prompt History Collection and Dialogue Gradient Computation}
\label{fig:dialogue_gradient}
\end{figure}



In this section, we detail the process for constructing dialogue gradients, denoted as $\nabla \mathcal{D}$, utilizing the prefix prompt history $H$ obtained in the previous phase. The overall process is shown in Figure \ref{fig:dialogue_gradient}(b).

For optimizing API-based LLMs, it is essential to format the training data into a dialogue history structure, consisting of user message-assistant message pairs. The API-based LLM is optimized to predict the assistant's message given the user's message by language modeling objective. To this end, we introduce the dialogue gradient $\nabla \mathcal{D}$ as an objective-injected dialogue history prepared for optimization within API-based LLMs.

The core strategy for injecting the optimization objective into the dialogue history involves a sliding window of size $w$ across the prefix prompt history $H$. This window segments the history into parts, with the section within the window acting as the user message, and the subsequent history entry serving as the assistant message. This method effectively incorporates the optimization goal—seeking the next prefix prompt to enhance the metric score—into the dialogue history. The formalization of this methodology is presented below:

Given the prefix prompt history $H = \{(s_0, m_0), (s_1, m_1), \ldots, (s_k, m_k)\}$, and employing a sliding window of size $w$ across $H$, we define the dialogue gradient at each window position $i$, where $0 \leq i < k-w$, as follows:
\begin{equation}
\nabla \mathcal{D}_i = \{\{(s_i, m_i), \ldots, (s_{i+w-1}, m_{i+w-1})\}, s_{i+w}\},
\end{equation}
where the tuples within the window represent the user messages and the immediate next prefix prompt outside the window serves as the assistant message. 

Finally, we enrich the dialogue gradients by appending critical data schemas including the dataset description $d$, an exemplar data example $E$, and the LLM prompt $p$. This enhanced dialogue gradient is then utilized to invoke the finetuning API for optimizing the API-based LLM.



\section{Experiments}

\subsection{Experiment Setup}

To evaluate the GPTA framework, we undertake comprehensive experiments across six benchmark datasets, spanning three critical domains in natural language processing. Each dataset is selected to represent a distinct task, allowing for a detailed assessment of the framework's capabilities:

\textbf{Machine Reading Comprehension:} This domain tests the model's ability to comprehend a given document and answer questions derived from it. We utilize the SQuAD dataset \citep{rajpurkar-etal-2016-squad} to examine the model's ability in sequence labeling, and the RACE dataset \citep{lai-etal-2017-race} to evaluate its semantic matching capability.

\textbf{Sentiment Analysis:} The focus is on the model's capacity to categorize sentences according to their sentiment. The Yelp \citep{asghar2016yelp} and Twitter \citep{giachanou2016like} datasets serve to assess the model's proficiency in multi-class and binary classification, respectively.

\textbf{Abstractive Summarization:} This domain challenges the model to distill a long document into a concise summary. To evaluate the model's performance on generative tasks, we selected the widely recognized CNN/Daily Mail \citep{hermann2015teaching} and XSum \citep{narayan2018don} datasets. 

To manage time and computational limits, we adopted random sampling for validation and testing across datasets, selecting 100 validation samples for preliminary assessments and 1,000 test samples for final evaluations. This method reliably mirrors full test set outcomes. For training, we sampled 10,000 and 100,000 instances from each training dataset to mimic low- and high-resource conditions, facilitating a broad evaluation. Implementation details are in Appendix \ref{app:implementation}.



\subsection{Experiment Results}


\begin{table}[t!]
\small
\centering
\scalebox{0.83}{
\begin{tabular}{l|cc|cc|cc|c}
\toprule
& \multicolumn{2}{c|}{\textbf{Sentiment Analysis}} & \multicolumn{2}{c|}{\textbf{Machine Reading }} & \multicolumn{2}{c|}{\textbf{Abstractive Sum}} & \multirow{2}{*}{\textbf{Avg Rank}} \\
\cmidrule(lr){2-3} \cmidrule(lr){4-5} \cmidrule(lr){6-7}
 \textbf{Models} & \textbf{Yelp (Acc)} & \textbf{Twitter (Acc)} & \textbf{SQuAD (F1)} & \textbf{RACE (Acc)} & \textbf{CNN/DM (R1)} & \textbf{XSum (R1)} & \\
\midrule
\textbf{Baseline} & 64.78 [3] & 84.32 [3] & 62.21 [3] & 45.04 [3] & 60.78 [3] & 53.22 [3] & 3.0\\
\textbf{GPTA-Data} & 66.23 [2] & 84.42 [2] & 66.25 [2] & \textbf{49.20} [1] & 61.86 [2] & \textbf{54.65} [1] & 1.6 \\
\textbf{GPTA} & \textbf{67.75} [1] & \textbf{84.88} [1] & \textbf{67.68} [1] & 47.60 [2] & \textbf{62.15} [1] & 54.57 [2] & \textbf{1.3} \\
\midrule
\rowcolor{gray!10}
\multicolumn{8}{c}{\textbf{\textsl{Low Resource Setting (10,000 training data)}}} \\
\midrule
\textbf{Baseline} & 61.22 [3] & 80.87 [3] & 45.34 [3] & 33.21 [3] & 52.35 [3] & 42.44 [3] & 3.0 \\
\textbf{GPTA-Data} & 61.40 [2] & 81.42 [2] & 46.34 [2] & \textbf{36.21} [1] & \textbf{53.50} [1] & \textbf{42.78} [1] & \textbf{1.5} \\
\textbf{GPTA} & \textbf{61.53} [1] & \textbf{81.70} [1] & \textbf{47.86} [1] & 35.33 [2] & 52.90 [2] & 42.61 [2] & \textbf{1.5} \\
\bottomrule
\end{tabular}
}

\caption{Experimental results of performance across training scenarios. ``Baseline" is the model optimized via conventional gradient descent.``GPTA" represents models trained with the GPTA framework, and ``GPTA-Data" includes models with three example data points in LLM prompts. Dataset names are followed by metrics: ``Acc" for accuracy, ``F1" for F1-score, and ``R1" for ROUGE-1 F-score \citep{lin2003automatic}. In each column, the number in [$\cdot$] indicates the ranking per setting. Following \cite{touvron2023llama}, ``Avg Rank" calculates the average of all rankings, indicating overall natural language understanding capability.}
\label{tab:main_exp}
\end{table}

\begin{figure}[!t]
\begin{center}
\includegraphics[width=0.93\textwidth]{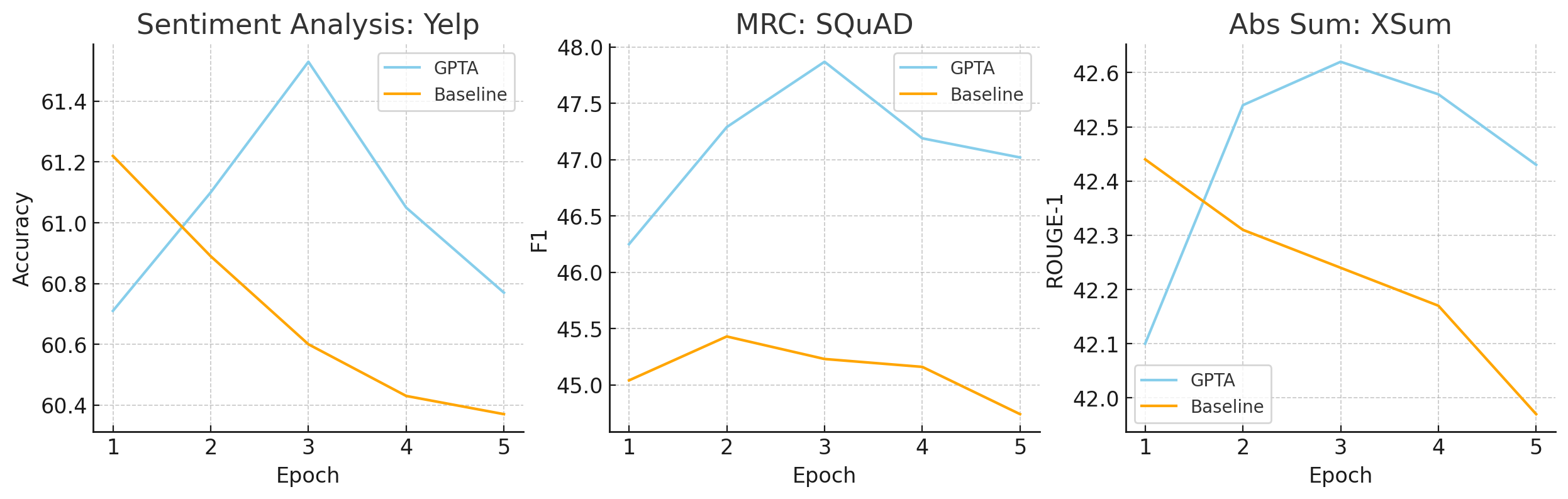}
\end{center}
\caption{The performance evaluation of GPTA on low-resource training setting over epochs.}
\label{fig:low_res}
\end{figure}

The experimental results presented in Table \ref{tab:main_exp} demonstrate the effectiveness of the GPTA framework across different training settings and domains. We highlight several critical findings:


\textbf{GPTA significantly enhances downstream task models in both high and low resource settings.} Analysis within each dataset segment shows GPTA's consistent superiority over the baseline model in all tasks, under both standard and low resource conditions. Specifically, GPTA achieves up to a 5.47-point enhancement in standard settings and a 3.00-point increase in low-resource settings over the baseline. These findings underscore GPTA's efficient LLM utilization to boost learning outcomes, regardless of resource levels.


\textbf{GPTA effectively addresses overfitting in low-resource settings.} Despite smaller absolute gains in low-resource setting, GPTA exhibits stronger overfitting resistance. As shown in Figure \ref{fig:low_res}, the baseline performance drops after the initial epoch, while GPTA demonstrates progressive improvement, overtaking baseline performance without experiencing severe performance drops. This resilience is invaluable for data-scarce NLP tasks, positioning GPTA as a dependable solution for maintaining steady model performance. The evaluation on all six datasets can be found in Figure \ref{fig:low_res_full} in Appendix \ref{app:aux_figures}.

\textbf{GPTA operates effectively without direct exposing data to the LLM.} The comparative performance of the GPTA and GPTA-Data configurations demonstrates that the GPTA framework can achieve competitive or better outcomes without incorporating exemplar data into the LLM prompts. This highlights GPTA's capability to enhance the performance of downstream models but avoid the need for direct data input to LLMs, thereby reducing potential privacy and security risks.

\textbf{GPTA demonstrates superior performance in discriminative tasks compared to generative tasks.} Task-specific analyses reveal the GPTA framework's proficiency in discriminative domains, like sentiment analysis and machine reading comprehension, as opposed to generative domains like abstractive summarization.  Notably, even within the generative tasks, the framework shows a greater improvement on the CNN/DM dataset which is designed more toward to extractive tasks, further validates its relative strength in tasks with discriminative characteristics.

\section{Analysis}


\subsection{Optimal Data Quantity for LLM Optimization}
\begin{table}[t!]
\small
\centering
\scalebox{1}{
\begin{tabular}{l|c c c}
\toprule
\textbf{\# Data} & \textbf{Yelp (Acc)} & \textbf{SQuAD (F1)} & \textbf{CNN/DM (R1)} \\
\midrule
\textbf{50} & \textbf{67.75} & \textbf{67.68} & 62.11 \\
\textbf{100} & 65.43 & 62.96 & \textbf{62.23} \\
\textbf{150} & 66.52 & 66.04 & 61.82 \\
\textbf{200} & 65.47 & 64.85 & 61.40 \\
\bottomrule
\end{tabular}
}
\caption{Experimental results for sentiment analysis, machine reading comprehension, and abstractive summarization at different scales of LLM optimizing data.}
\label{tab:finetune_data_amount}
\end{table}

\label{sec:llm_data}

To identify the optimal training data volume for dialogue gradient computation, we selected a benchmark dataset from each domain, varying data volumes during training. As detailed in Table \ref{tab:finetune_data_amount}, the peak performance for discriminative tasks (Yelp and SQuAD) occurs with 50 data points, while abstractive summarization tasks (CNN/DM) require 100 data points. Performance uniformly drops across domains beyond 150 data points, corroborating the API provider's recommendations and suggesting that surpassing this limit diminishes the LLM's generalization ability.




\subsection{LLM Prefix Prompt Searching Accuracy}
\label{sec:llm_training}



\begin{figure}[!t]
\begin{center}
\includegraphics[width=0.92\textwidth]{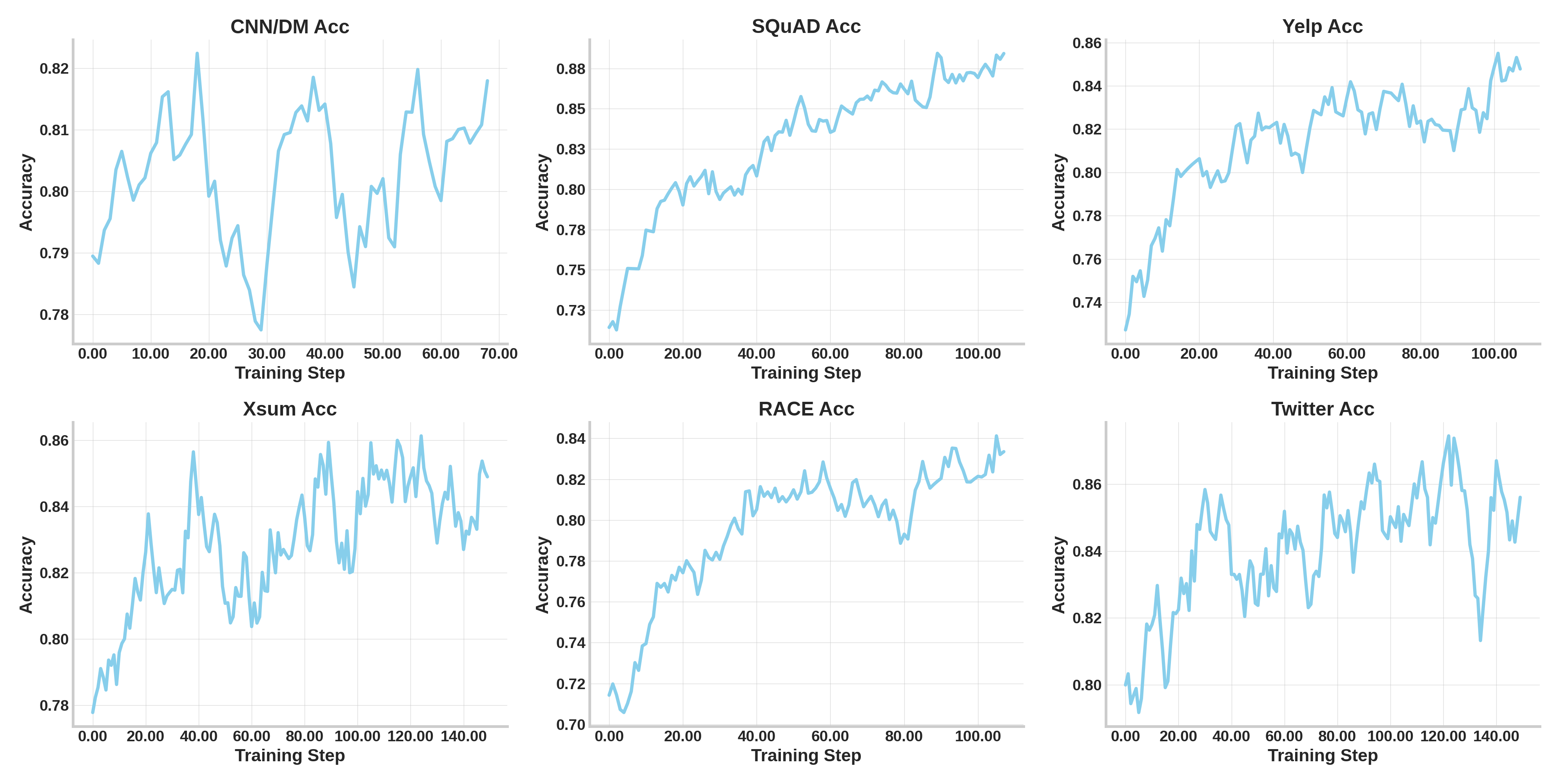}
\end{center}
\caption{The training accuracy of LLM finding the next prefix prompt enhancing the downstream task model performance. The performance at step 0 is the performance of the \textit{gpt-3.5turbo-0613}.}
\label{fig:llm_train_acc}
\end{figure}

Our analysis of the optimized TA LLM demonstrates its improved capability to accurately generate subsequent prefix prompts during the optimization phase, as evidenced in Figure \ref{fig:llm_train_acc}. By employing dialogue gradients for optimization, the LLM's prefix generation accuracy across all six datasets rose by up to 15\%. This enhancement validates the effectiveness of dialogue gradients in refining LLMs for specific tasks.

Furthermore, a strong correlation exists between the improved prefix prompt generation accuracy and the downstream task model's performance improvements shown in Table \ref{tab:main_exp}. This highlights that the primary factor in boosting downstream model performance is its dynamic interaction with the LLM through high-quality prefix prompts.


\subsection{Prefix Prompt Analysis}
\label{sec:case}

\begin{table}[t!]
\centering
\small
\scalebox{0.72}{
\begin{tabular}{l|lll}
\toprule
\textbf{Dataset} & \textbf{Prompt \#1} & \textbf{Prompt \#2} & \textbf{ Prompt \#3} \\
\midrule
\textbf{Yelp} & Check out trusted reviews & Discover popular attractions & Find honest business information \\
\textbf{Twitter} & Emotion Insights & Exploring Emotional Tweets & Understanding Emotions \\
\midrule
\textbf{SQuAD} & Addressing Unanswerable Questions & Improving Paragraph Understanding & Enhancing Answer Verification \\
\textbf{RACE} & Designing informative passages & Enhancing dataset diversity & Exploring contextual understanding \\
\midrule
\textbf{CNN/DM} & Sift through crucial content & Delve into valuable content & Abstract significant content \\
\textbf{XSum} & Exploring importance of context & Studying role of discourse structure & Analyzing role of sentence compression \\
\bottomrule
\end{tabular}
}
\caption{The top-3 prefix prompts for each dataset.}
\label{tab:soft_prefix_strategies}
\end{table}


When we investigate the top-3 prefix prompts showcased in Table \ref{tab:soft_prefix_strategies}, an intriguing pattern emerged: these prompts tend to align more closely with the domain they pertain to rather than the intricacies of individual datasets. This pattern indicates that the GPTA framework is adept at identifying prefix prompts that are not just dataset-specific but have broader applicability across a given domain. Such a characteristic is highly beneficial as it suggests that prefix prompts optimized for one dataset might be effectively utilized to enhance model performance on other datasets within the same domain, offering a strategy for cross-dataset performance improvement. Moreover, the analysis highlights that optimal prefix prompts, despite being five words or fewer, significantly enhance model performance, suggesting the most effective prompts succinctly capture a domain's essence.


\section{Conclusion}
In this paper, we introduce GPTA, a framework that uses Large Language Models to enhance downstream model training while addressing data security and legal challenges. By using LLMs to generate dynamic prefix prompts from dataset descriptions, GPTA improves NLP task performance through innovative training and optimization techniques. Tested across six datasets, GPTA has been shown to increase accuracy and reduce overfitting, particularly in data-scarce environments. Our work marks the first attempt at enhancing downstream model training with API-based LLMs through joint training, suggesting a potential path forward for developing more robust, domain-specific models.

\bibliography{colm2024_conference,anthology,custom}
\bibliographystyle{colm2024_conference}

\appendix
\newpage
\section{Appendix}
\subsection{Implementation Details}
\label{app:implementation}
To ensure a rigorous evaluation of the proposed framework's effectiveness, we deliberately simplify the selection of downstream task models. This approach aim to reduce their potential impact on the outcomes. Specifically, for tasks categorized under machine reading comprehension and sentiment analysis, we utilize the \textit{bert-base-uncased} model, chosen for its established baseline performance. For tasks related to abstractive summarization, the \textit{bart-base} model is selected, capitalizing on its proficiency in sequence-to-sequence text generation. Across all tasks, the \textit{gpt-3.5turbo-0613} model functions as the TA model, providing a uniform framework for performance assessment across varied domains.

During the training of downstream task models, we configure the learning rate $\alpha$ to $2e^{-5}$, incorporating a weight decay of $0.01$ through the use of the AdamW optimizer \citep{loshchilov2017decoupled}. In the phase of collecting soft prompt history, we iteratively prompt the LLM to accumulate a set of $k=50$ soft prompt histories, setting the temperature to $1.0$. For the computation of dialogue gradients, a sliding window of size $w=5$ is employed. Each task undergoes an alternating training cycle between the downstream task model and the LLM, as detailed in Section \ref{sec:method}, spanning $5$ epochs. The checkpoint showcasing the best performance is subsequently reloaded for further analysis.

\subsection{LLM Prompts}
\label{app: llm_prompts}
\begin{table*}[!th]
\centering
\begin{tabular}{|>{\raggedright\arraybackslash}p{0.45\textwidth}|>{\raggedright\arraybackslash}p{0.45\textwidth}|}
\hline
\textbf{System Prompt} & \textbf{User Prompt} \\
\hline
\vspace{2pt}
You are an prefix generation model. The prefix you generated is used to help Sentiment Analysis model to better distinguish. \newline

The model is based on BERT, so the prefix will be added as [CLS]prefix + context. The prefix should be a short sentence.
Please only output the prefix. \newline

 &
\vspace{1pt}
\textbf{Dataset Description $d$:} The description of the dataset is as follows: The Yelp reviews dataset consists of reviews from Yelp. It is extracted from the Yelp Dataset Challenge 2015 data. \newline \newline
\textbf{Optional Exemplar Data $E$:} [if showdata is true, examples from the dataset are provided for reference.] \newline \newline
\textbf{Prefix Prompt History $H$: }This is the history of prefixes and their corresponding accuracy scores that you generated previously in ascending order: \newline
\texttt{<HISTORIES>} \newline \newline
\textbf{LLM Prompt $p$:} Please follow the pattern of history and generate \{prefixes\_num\} new prefixes that could potentially improve the accuracy score. The prefix should be a short sentence. The prefix could be related to the dataset, or it could be a general sentence. \newline
Please only output the prefixes as a json.
Example: \{"prefixes": ["prefix1", "prefix2", ...]\}
\vspace{5pt}
\\
\hline
\end{tabular}
\caption{System and user prompts for prefix generation for Yelp Sentiment Analysis benchmark dataset.}
\label{tab:prefix_generation}
\end{table*}

\subsection{Auxiliary Figures}
\label{app:aux_figures}
\begin{figure}[!th]
\begin{center}
\includegraphics[width=0.85\textwidth]{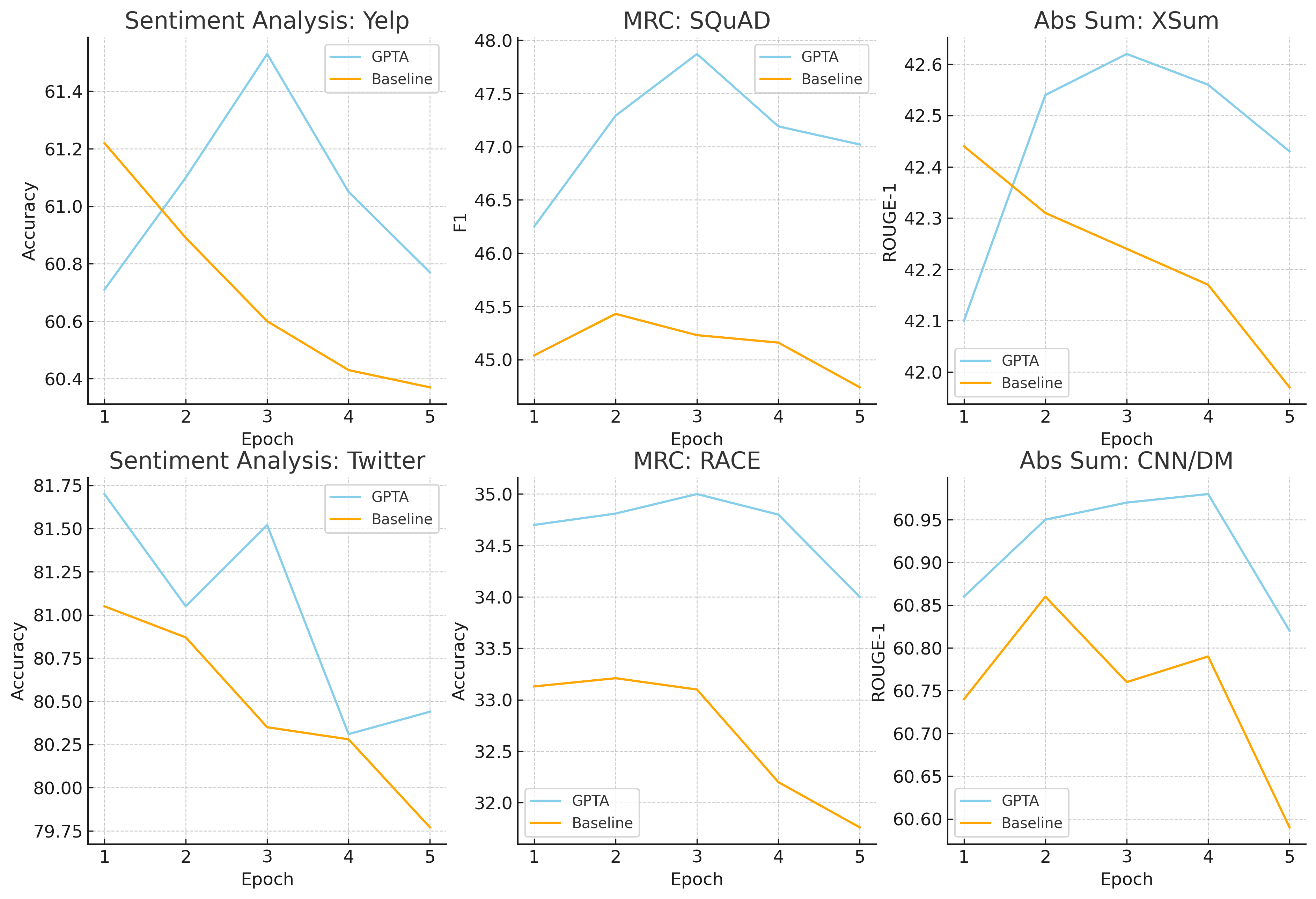}
\end{center}
\caption{The performance evaluation of GPTA on low-resource training setting over epochs on all datasets.}
\label{fig:low_res_full}
\end{figure}

\begin{figure}[!th]
\begin{center}
\includegraphics[width=0.92\textwidth]{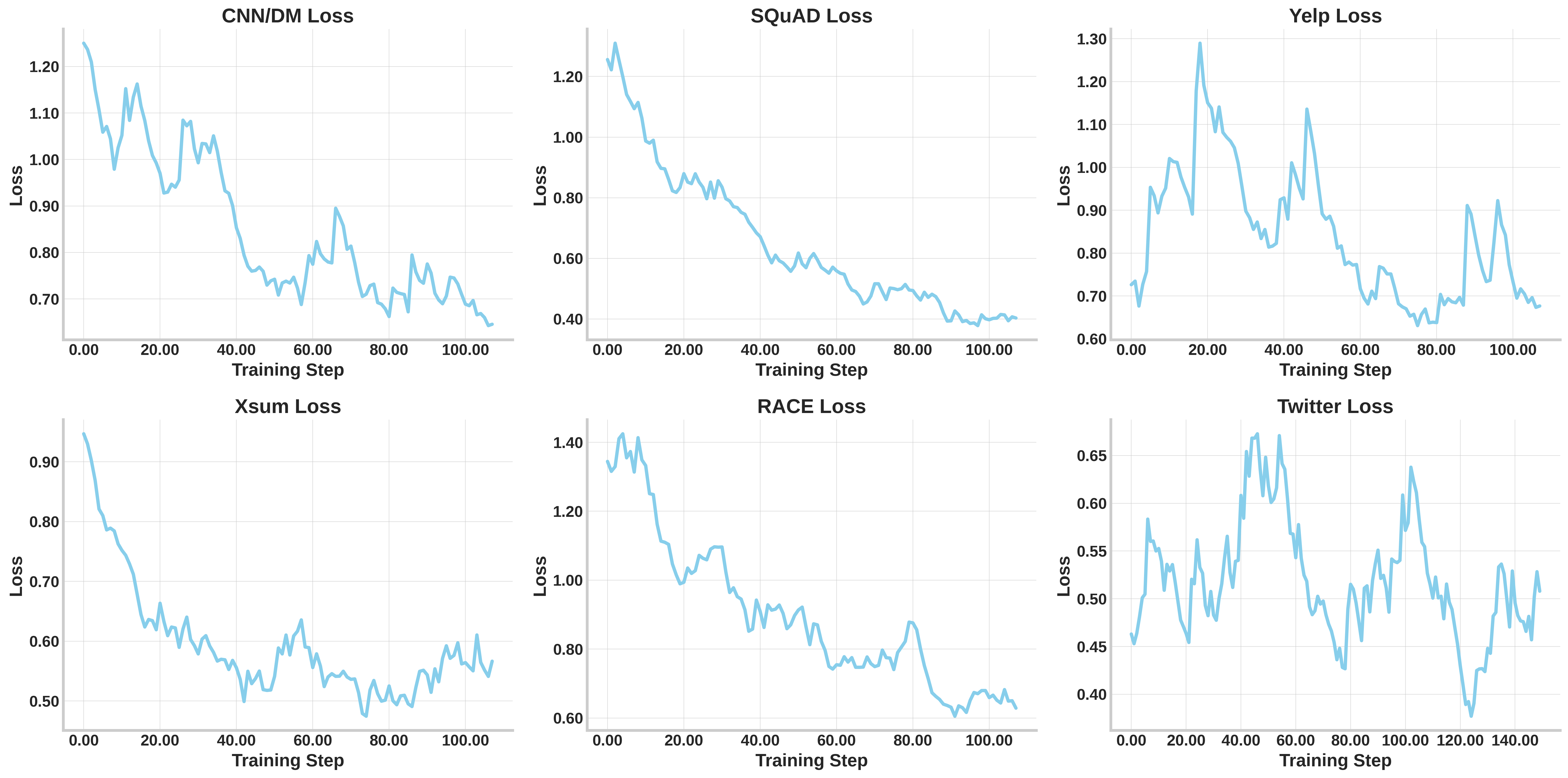}
\end{center}
\caption{The LLM optimization loss on all six datasets.}
\label{fig:llm_train_loss}
\end{figure}
\end{document}